\documentclass[a4paper,twoside]{article}

\usepackage[utf8]{inputenc}
\usepackage[T1]{fontenc}

\usepackage{epsfig}
\usepackage{subcaption}
\usepackage{xcolor}
\usepackage{calc}
\usepackage{amssymb}
\usepackage{amstext}
\usepackage{amsmath}
\usepackage{amsthm}
\usepackage{multicol}
\usepackage{pslatex}
\usepackage{apalike}
\usepackage{algorithm2e}
\usepackage[bottom]{footmisc}
\usepackage{SCITEPRESS}     % Please add other packages that you may need BEFORE the SCITEPRESS.sty package.

\usepackage{eso-pic}
\usepackage{graphicx} % Necesario para rotar el texto
\AddToShipoutPictureBG{%
  \AtPageLowerLeft{%
    \put(\LenToUnit{1.5cm},\LenToUnit{0.5\paperheight}){%
      \rotatebox{90}{\makebox[0pt][c]{\textcolor{gray}{\large \textbf{Preprint accepted for publication at the 21st International Conference on Computer Vision Theory and Applications (VISAPP 2026)}}}}%
    }%
  }%
}
\begin{document}

\title{Computing a Characteristic Orientation for Rotation-Independent Image Analysis}

% \author{\authorname{Cristian Valero-Abundio\sup{1}\orcidAuthor{0009-0007-1035-942X}, Emilio Sansano-Sansano\sup{1}\orcidAuthor{0000-0002-7814-0383}, Raúl Montoliu\sup{1}\orcidAuthor{0000-0002-8467-391X} and Marina Martínez García\sup{2}\orcidAuthor{0000-0002-2228-4396}}
\author{\authorname{Cristian Valero-Abundio\sup{1}, Emilio Sansano-Sansano\sup{1}, Raúl Montoliu\sup{1} and \\Marina Martínez García\sup{2}}
\affiliation{\sup{1}Institute of New Imaging Technologies, Universitat Jaume I, 12071 Castellón, Spain}
\affiliation{\sup{2}Mathematics Department, Universitat Jaume I, 12071 Castellón, Spain}
\email{\{cvalero, esansano, martigar, montoliu\}@uji.es}
}


\abstract{Handling geometric transformations, particularly rotations, remains a challenge in deep learning for computer vision. Standard neural networks lack inherent rotation invariance and typically rely on data augmentation or architectural modifications to improve robustness. Although effective, these approaches increase computational demands, require specialised implementations, or alter network structures, limiting their applicability. This paper introduces \textit{General Intensity Direction} (GID), a preprocessing method that improves rotation robustness without modifying the network architecture. The method estimates a global orientation for each image and aligns it to a canonical reference frame, allowing standard models to process inputs more consistently across different rotations. Unlike moment-based approaches that extract invariant descriptors, this method directly transforms the image while preserving spatial structure, making it compatible with convolutional networks. Experimental evaluation on the rotated MNIST dataset  shows that the proposed method achieves higher accuracy than state-of-the-art rotation-invariant architectures. Additional experiments on the CIFAR-10 dataset, confirm that the method remains effective under more complex conditions.}

\onecolumn \maketitle \normalsize \setcounter{footnote}{0} \vfill

\section{\uppercase{Introduction}}
\label{sec:introduction}

Deep learning (DL) has driven significant advances in computer vision, enabling improvements in image classification, object detection, and pattern recognition. However, neural networks often struggle with geometric transformations, particularly rotations. Standard architectures are not inherently rotation-invariant and typically require extensive data augmentation or complex architectural modifications to handle rotated inputs. While various solutions have been proposed, many introduce additional complexity, limiting their practicality for broader applications and making deployment more demanding in real-world pipelines.

Data augmentation is a widely used strategy to improve rotation robustness, but it significantly increases training time and does not guarantee full invariance. Models must learn redundant features from multiple orientations, leading to greater computational demands and larger effective training sets. An alternative is to design architectures that explicitly encode rotation invariance or equivariance, but these approaches often require specialised implementations and retraining from scratch, which can be difficult to integrate into existing workflows. A more practical solution is to improve rotation handling without modifying the underlying model. This allows pre-trained networks to be reused with minimal fine-tuning, reducing computational costs while maintaining compatibility with existing architectures and training procedures.

% Several works have addressed rotation invariance through different strategies, including \textit{moment invariants}, architectural modifications, and preprocessing techniques. Moment-based approaches, such as Hu~\cite{hu1962} and Zernike~\cite{khotanzad1990} moments, compute descriptors that remain unchanged under geometric transformations, capturing the intrinsic orientation of an image without requiring explicit alignment. While effective for feature extraction, these descriptors do not produce a transformed image with a standardised orientation and may not be ideal for neural networks that rely on structured spatial information. Some methods attempt to estimate a principal axis of rotation from second-order central moments or gradient distributions and rotate the image accordingly~\cite{flusser2000}, but such approaches often involve complex computations and can be sensitive to noise or background variations, which reduces their robustness in practice.

Various strategies address rotation invariance, including \textit{moment invariants}, architectural modifications, and preprocessing. Moment-based methods, such as Hu~\cite{hu1962} and Zernike~\cite{khotanzad1990}, compute descriptors that capture orientation without explicit alignment. While effective for feature extraction, they do not produce standardized images, making them suboptimal for neural networks relying on spatial structure. Other approaches estimate a principal axis from central moments or gradients to rotate the image accordingly~\cite{flusser2000}. However, these often involve complex computations and exhibit sensitivity to noise or background variations, reducing their practical robustness.

Other strategies involve modifying the architecture to enforce equivariance or incorporating trainable mechanisms that adapt feature extraction to input orientation~\cite{cohen2016,dieleman2015,worrall2017,mo2024,hao2022}. While these techniques can be highly effective, they increase model complexity and reduce compatibility with standard architectures, which complicates reuse of existing models. Data augmentation remains the most widely used approach, but it demands extensive training and does not always generalise well to unseen transformations. Preprocessing methods, by contrast, offer a simpler alternative that allows existing models to handle rotations without modification.

% This paper introduces \textit{General Intensity Direction} (GID), a preprocessing method that enhances rotation invariance while maintaining compatibility with existing artificial neural networks (ANNs). The method estimates a global orientation for each input image and rotates it to a canonical alignment, transforming the image itself rather than extracting invariant descriptors. This ensures that the spatial structure remains consistent for downstream processing, making it suitable for convolutional networks without requiring architectural modifications. GID can be applied as a standalone preprocessing step or integrated before the first convolutional layer. Experimental results on rotated image datasets demonstrate substantial accuracy improvements compared to state-of-the-art approaches, highlighting its practical value as a lightweight alternative to more complex rotation-invariant architectures and its ease of integration into typical training and inference setups.

This paper introduces \textit{General Intensity Direction} (GID) , a preprocessing method enhancing rotation invariance while maintaining compatibility with existing neural networks. The method estimates a global orientation for each image and rotates it to a canonical alignment, transforming the image rather than extracting invariant descriptors. This preserves spatial structure for downstream processing, ensuring suitability for convolutional networks without architectural modifications. GID can function as standalone preprocessing or be integrated before the first convolutional layer. Experimental results on rotated datasets show accuracy improvements over state-of-the-art approaches , highlighting GID as a lightweight alternative to complex architectures and its ease of integration into standard training and inference setups.

% The remainder of the paper is organised as follows. Section~\ref{sec:related_work} presents related work, summarising existing approaches to rotation invariance and their limitations. Section~\ref{sec:proposed_method} describes the proposed preprocessing method and its integration into neural networks, detailing how it can be used within standard pipelines. Section~\ref{sec:experiments_results} details the experimental setup, datasets, and performance comparison with state-of-the-art methods. Section~\ref{sec:conclusions} concludes with a summary of the findings and potential directions for future research.

\section{\uppercase{Related work}}
\label{sec:related_work}

%Improving rotation invariance in neural networks has been widely studied, with various approaches trying to address this challenge. One of the most common strategies is data augmentation, in which images are artificially rotated to expose the model to multiple orientations during training. Although simple and easy to implement, this method significantly increases training time and does not guarantee complete invariance. Since neural networks must still learn redundant patterns from samples rotated differently, the capacity of the model is used inefficiently, and generalisation to unseen orientations remains limited \cite{quiroga2019,vandyk2001,simard2003}. In addition, augmentation requires a substantial increase in training resources, including larger datasets, longer training times, and higher memory consumption, as the network must process multiple transformed versions of the same input. This inefficiency is particularly problematic for large-scale applications or real-time systems, where computational costs and latency must be minimised. As a result, data augmentation does not provide theoretical guarantees of invariance, meaning that the network may still struggle with rotations beyond the range seen during training, leading to performance degradation in unconstrained settings.

Improving rotation invariance in neural networks is a widely studied challenge addressed through various methodologies. A common strategy is data augmentation, where images are artificially rotated to expose the model to multiple orientations during training. While simple to implement, this method increases training time and does not guarantee complete invariance. Since networks must learn redundant patterns from differently rotated samples, model capacity is used inefficiently and generalization to unseen orientations remains limited \cite{quiroga2019,vandyk2001,simard2003}. Additionally, augmentation requires substantial resources—larger datasets, longer training, and higher memory—to process multiple transformed versions of the same input. This inefficiency is particularly problematic for large-scale or real-time systems where computational costs and latency must be minimized. Ultimately, data augmentation lacks theoretical guarantees, potentially leading to performance degradation in unconstrained settings when encountering rotations beyond the training range.

%Another category of methods consists of modifying the network architecture to explicitly encode rotation equivariance. These approaches leverage \textit{group convolutions} \cite{cohen2016} and enforce weight sharing for rotations using symmetry-aware layers to maintain equivariance at multiple levels of the model \cite{fasel2006,dieleman2015,gens2014}. By preserving transformation properties throughout the network, robustness is improved without requiring excessive data augmentation. However, the main drawback of these methods is that they often require significant architectural modifications, making them less compatible with pre-trained models and standard deep learning frameworks. Additionally, the computational overhead associated with handling larger symmetry groups can make deployment more expensive.

Another category of methods modifies network architectures to explicitly encode rotation equivariance. These approaches utilize \textit{group convolutions} \cite{cohen2016} and enforce weight sharing via symmetry-aware layers to maintain equivariance across model levels \cite{fasel2006,dieleman2015,gens2014}. Preserving transformation properties improves robustness without requiring excessive data augmentation. However, these methods often demand significant architectural changes, limiting compatibility with pre-trained models and standard frameworks. Furthermore, the computational overhead associated with larger symmetry groups can increase deployment costs.

To bypass the limitations of augmentation and equivariant architectures, trainable feature extraction mechanisms have been developed. These include spatial transformer networks \cite{jaderberg2015}, adaptive pooling \cite{laptev2016}—adjusting representations based on orientation—steerable filters \cite{freeman1991,worrall2017,delchevalerie2021,jenner2022,dai2017,mo2024}, and rotating filters \cite{orn2017,marcos2017}. While flexible, these methods increase model complexity, often requiring extra supervision or tuning, and their performance relies on the quality of learned transformations. Furthermore, they may still struggle with extreme rotations or distortions underrepresented in the training data.

% A different line of research focuses on invariant feature representations \cite{lowe1999}, where inputs are mapped into a canonical space that is unaffected by rotations \cite{rublee2011,hao2022}. This can be achieved through handcrafted transformations \cite{lowe1999} or feature aggregation \cite{yu2015}. Such methods effectively eliminate the need for rotation-specific training, but they often introduce a loss of spatial information. In this respect, ensuring that the transformation preserves all relevant features of the input while achieving full invariance remains a challenge.

Other research focuses on invariant feature representations \cite{lowe1999}, where inputs are mapped into a canonical space unaffected by rotations \cite{rublee2011,hao2022}. This is achieved through handcrafted transformations \cite{lowe1999} or feature aggregation \cite{yu2015}. While these methods eliminate rotation-specific training, they often introduce a loss of spatial information. Consequently, ensuring that transformations preserve all relevant features while achieving full invariance remains a challenge.

%A well-established class of methods that relate to our work is \textit{moment invariants}, which extract rotation-independent descriptors from an image. Classical techniques such as Hu \cite{hu1962} moments and Zernike \cite{khotanzad1990} moments provide feature representations that remain unchanged under rotation, enabling robust classification. More recent extensions apply higher-order moments to estimate dominant orientations and align images accordingly \cite{flusser2000}. These methods share similarities with our approach in that they aim to standardise orientation, but they generally do not reconstruct the image in a canonical form. Instead, they produce invariant descriptors that are often decoupled from spatial structure, which may be suboptimal for convolutional neural networks that rely on spatially ordered features.

A related class of methods is \textit{moment invariants}, which extract rotation-independent descriptors. Classical techniques like Hu \cite{hu1962} and Zernike \cite{khotanzad1990} moments provide feature representations unchanged under rotation for robust classification. Recent extensions use higher-order moments to estimate dominant orientations and align images \cite{flusser2000}. While similar to our approach in seeking to standardize orientation, these methods generally do not reconstruct the image in a canonical form. Instead, they produce invariant descriptors often decoupled from spatial structure, which can be suboptimal for convolutional neural networks relying on spatially ordered features.

% Despite these advancements, existing approaches either require substantial computational resources, extensive retraining, or introduce architectural constraints that limit their applicability. A preprocessing-based solution offers a simpler alternative by transforming inputs before they are processed by a standard neural network. Our proposed approach enables rotation invariance without modifying network architectures, making it compatible with pre-trained models. It determines a global orientation for each image and rotates it accordingly to align it with a canonical reference. Unlike moment-based methods that produce invariant descriptors, our method outputs a transformed image that preserves spatial relationships, ensuring consistency in subsequent neural network processing. This approach provides a lightweight alternative to architectural modifications while maintaining interpretability and computational efficiency.

Despite these advancements, existing approaches often require substantial resources, extensive retraining, or architectural constraints. A preprocessing-based solution offers a simpler alternative, transforming inputs before standard neural network processing. Our approach enables rotation invariance without architectural modifications, ensuring compatibility with pre-trained models. It determines a global orientation for each image, rotating it to align with a canonical reference. Unlike moment-based descriptors, our method outputs a transformed image that preserves spatial relationships and ensures consistency for subsequent processing. This provides a lightweight alternative to architectural changes while maintaining interpretability and efficiency.

\section{\uppercase{Proposed Method}}
\label{sec:proposed_method}

The goal of the proposed \textit{General Intensity Direction} (GID) method is to determine a global angle for each image and rotate it accordingly, using the image centre as the pivot point. %An illustrative video demonstrating the rotation process is available as supplementary material \url{https://youtu.be/KESPYngSe6c}.

For clarity, we first describe the procedure for a greyscale image, which naturally extends to cases with multiple channels.

Let $I$ be an image of size $H \times W$, where the intensity at each pixel is denoted by $I(i,j):\Omega \to [0 , 1]$. The discrete spatial domain of a single-channel image is given by:

\begin{equation}
    \Omega = \{(i,j) \mid 0 \leq i < H,\, 0 \leq j < W,\, i, j \in  \mathbb{N} \} \subset \mathbb{N}^2
\end{equation}

The method operates on the original pixel coordinates, applying a rotation to them before retrieving the corresponding intensity values. However, when $\alpha \neq k\pi/2$ for some integer $k$, the rotation does not align perfectly with the discrete grid, requiring interpolation. The procedure consists of the following steps:

\begin{enumerate}
    \item Since the rotation is performed around the image center, we first shift the coordinate system so that the origin is at the image center:
    \begin{equation}
        (c_y, c_x) = \Bigg( \frac{H - 1}{2},  \frac{W - 1}{2} \Bigg)
    \end{equation}
    The transformed coordinate domain is then:
    \begin{equation}
        \Omega_c = \{ (i , j) - (c_y, c_x) \mid (i, j) \in \Omega \} \subset \mathbb{Q}^2
    \end{equation}

    \item The next step is to compute the orientation of each pixel relative to the new coordinate system $\Omega_c$:
    \begin{equation}
        \theta_{i,j} = \operatorname{atan2}(i, j), \quad \forall (i, j) \in \Omega_c,
    \end{equation}
    where $\theta_{i,j}$ lies in the range $[-\pi, \pi)$. 

    \item The global angle is then computed by accumulating the sines and cosines of the local angles, weighted by their corresponding pixel intensities:
    \begin{align*}
        S_{\sin} 
        &= \sum_{(i,j) \in \Omega_c} I(i + c_y, j + c_x) \sin\bigl(\theta_{i,j}\bigr), \\
        S_{\cos}
        &= \sum_{(i,j) \in \Omega_c} I(i + c_y, j + c_x) \cos\bigl(\theta_{i,j}\bigr), \\
        \alpha 
        &= \operatorname{atan2}(S_{\sin}, S_{\cos}), \quad \alpha \in [-\pi, \pi)
    \end{align*}

    \item Finally, we rotate the coordinates in $\Omega_c$ by the computed angle $\alpha$ and shift them back to the original domain $\Omega$:
    \begin{equation}
        \Omega_{\text{rot}, \alpha} = 
        \begin{pmatrix}
            \cos(\alpha) & -\sin(\alpha) \\[6pt]
            \sin(\alpha) & \;\;\cos(\alpha)
        \end{pmatrix}
        \begin{pmatrix}
            i \\[2pt]
            j
        \end{pmatrix}
        + 
        \begin{pmatrix}
            c_y \\[2pt]
            c_x
        \end{pmatrix} \subset \mathbb{R}^{2}
    \end{equation}
\end{enumerate}

We apply this transformation to each pixel in $I$, using a standard interpolation scheme (e.g., nearest-neighbour or bilinear interpolation) to resample intensity values, and fill out-of-bound regions with zeros. Denoting the transformed image as $I_{\text{rot}}$, we obtain a representation that is invariant from the original orientation of $I$.

Although the derivation assumes a single channel image for clarity, the method naturally extends to multichannel data $I:\Omega \to [0,1]^m$. This can be achieved either by computing a single global angle $\alpha$ from an aggregate measure across all channels, or by assigning each channel its own angle when application-specific considerations justify it. In both cases, the rotation is applied consistently across all channels to maintain spatial coherence.

\section{\uppercase{Experiments and results}}
\label{sec:experiments_results}

\subsection{\uppercase{Datasets}}
\label{subsec:datasets}

% In this study, we employed the MNIST dataset \cite{lecun1998}, a widely used benchmark in the field of machine learning and computer vision. The MNIST dataset consists of a collection of 70,000 grayscale images of handwritten digits, ranging from 0 to 9. These images are divided into a training set of 60,000 samples and a test set of 10,000 samples. The labels in MNIST correspond to the digit represented in each image. Since the original MNIST images are well-centred and naturally aligned, it serves as a baseline for evaluating how the preprocessing method performs under standard conditions. We apply a padding transformation to the original $28 \times 28$ images, expanding them to $32 \times 32$ pixels.

In this study, we employed the MNIST dataset \cite{lecun1998}, a standard benchmark in machine learning and computer vision. It comprises 70,000 grayscale images of handwritten digits from 0 to 9, split into 60,000 training and 10,000 test samples. Since MNIST images are well-centered and naturally aligned, they provide a baseline for evaluating our preprocessing under standard conditions. Finally, the original $28 \times 28$ images are padded to $32 \times 32$ pixels.

% To compare our results with previous works on rotation-invariant models, we also used the RotMNIST dataset \cite{larochelle2007}. This variant contains the same digits as the original MNIST dataset but with random rotations applied to each image, introducing a greater level of variability. The rotated MNIST dataset is commonly used to assess a model’s ability to generalise to unseen orientations, making it suitable for evaluating the impact of our preprocessing method.

To compare our results with previous works on rotation-invariant models, we also utilized the RotMNIST dataset \cite{larochelle2007}. This variant applies random rotations to the original MNIST digits, increasing variability. As a standard benchmark for assessing generalization to unseen orientations, RotMNIST is well-suited for evaluating the impact of our preprocessing method.

% In addition, we used the CIFAR-10 dataset~\cite{krizhevsky2009} to further evaluate the robustness of the proposed method on more complex and natural images with background. CIFAR-10 consists of 60,000 colour images of size $32 \times 32$ pixels, distributed evenly across 10 different object classes such as airplanes, cars, birds, and animals. The dataset is split into 50,000 training samples and 10,000 test samples. Unlike MNIST, the images in CIFAR-10 contain natural backgrounds and higher variability in appearance, while still keeping the main object roughly centred. To ensure consistency across datasets, all images are converted to grayscale, reducing the number of channels to one. To align the input size with the requirements of our preprocessing method, we apply a padding operation that expands each image from $32 \times 32$ to $46 \times 46$ pixels.

In addition, we used the CIFAR-10 dataset~\cite{krizhevsky2009} to evaluate the robustness of the proposed method on more complex images with natural backgrounds. CIFAR-10 consists of 60,000 color images ($32 \times 32$ pixels) across 10 object classes, split into 50,000 training and 10,000 test samples. Unlike MNIST, these images feature natural backgrounds and higher appearance variability while keeping the main object roughly centered. To ensure consistency across datasets, all images are converted to grayscale. Finally, to align with our preprocessing requirements, a padding operation expands each image from $32 \times 32$ to $46 \times 46$ pixels.

\subsection{Models and Training Protocol}
\label{subsec:model_training}

% As our baseline, we employ Conv32, a custom convolutional neural network designed for image classification. The model consists of three convolutional layers with 32, 64, and 128 filters of size $3 \times 3$ (see Figure \ref{fig:cnn32_diagram}). Each convolutional layer is followed by batch normalization and a ReLU activation function. To progressively reduce spatial dimensions and computational cost, a $2 \times 2$ max pooling operation is applied after every convolutional layer. To enhance generalization, we introduce a dropout layer with a rate of 0.4 after the last convolutional block. The feature maps are then flattened and passed through a fully connected layer with 256 units and ReLU activation, followed by a second dropout layer (0.5) before the final classification layer, which outputs 10 neurons corresponding to the target classes. We follow the same training protocol as in previous works \cite{mo2024}. Specifically, the model is trained with a batch size of 100. The Adam optimizer is used with an initial learning rate of $10^{-4}$, which is reduced by a factor of 0.8 every ten epochs to stabilize convergence. No data augmentation techniques were applied during the training process.

\begin{figure*}[t!]
\centering
\includegraphics[width=320pt]{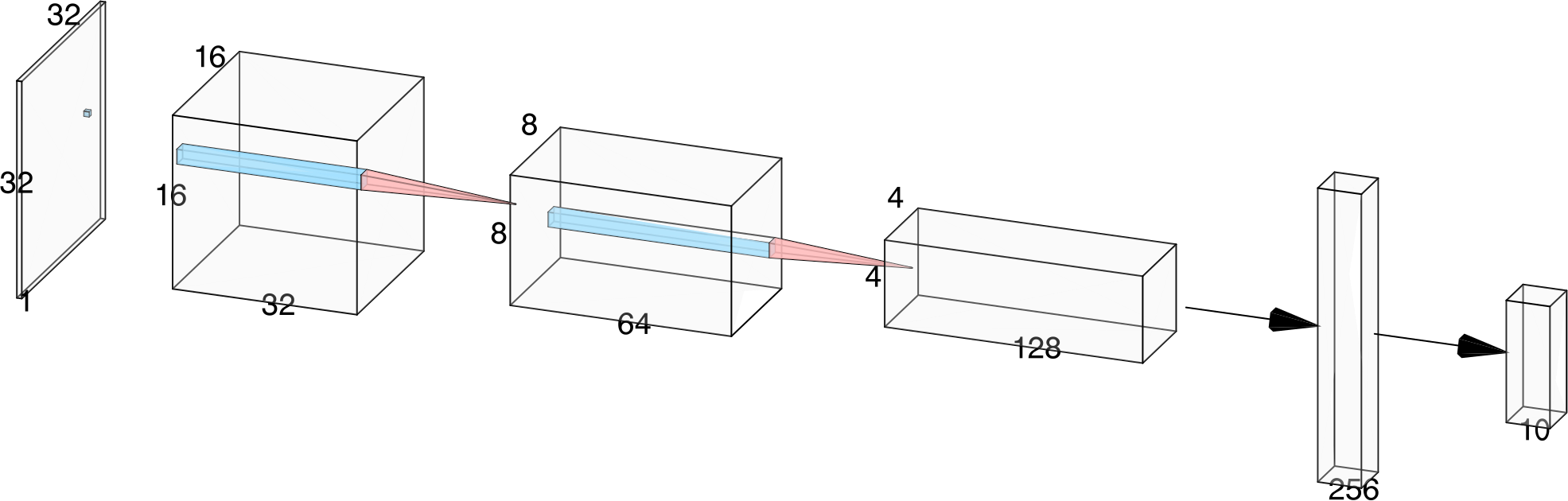}
\caption{\textit{Conv32} architecture diagram.}
\label{fig:cnn32_diagram}
\end{figure*}

As a baseline, we employ Conv32, a custom convolutional neural network for image classification. The model features three convolutional layers with 32, 64, and 128 filters of size $3 \times 3$ (Figure \ref{fig:cnn32_diagram}), each followed by batch normalization and ReLU activation. To reduce spatial dimensions, a $2 \times 2$ max pooling operation is applied after every convolutional layer. We include a dropout layer (0.4) after the last convolutional block to enhance generalization. Features are then flattened and passed through a fully connected layer (256 units, ReLU) and a second dropout (0.5) before the final 10-neuron classification layer. Training follows the protocol in \cite{mo2024}, using a batch size of 100 and the Adam optimizer. The initial learning rate of $10^{-4}$ is reduced by 0.8 every ten epochs. No data augmentation was applied during training.

For the experiments conducted on the CIFAR-10 dataset, we employ ResNet-18~\cite{he2016deep}, a residual network architecture adapted to better accommodate the input size of $46 \times 46$ pixels. The structure of the network remains unchanged from its standard implementation, including the number of convolutional layers, filter sizes, activation functions, and residual connections. The same training protocol described above is used for ResNet-18.

% It is important to note that the chosen architectures are intentionally simple and not specifically optimized for state-of-the-art performance. Their purpose is to provide a consistent and controlled setting in which to evaluate the effectiveness of the proposed preprocessing method. As such, the performance of the models should not be interpreted as a limitation of the method itself, which remains compatible with any network architecture.

It is important to note that the chosen architectures are intentionally simple and not optimized for state-of-the-art performance, providing a controlled setting to evaluate the preprocessing method’s effectiveness. Consequently, model performance should not be viewed as a limitation of the method itself, which remains compatible with any network architecture.

\subsection{Experiment 1: Robustness to Rotations}
\label{subsec:experiment1}

In this experiment, we assess the robustness of different models when subjected to rotated inputs. Specifically, Conv32 and ConvNet~\cite{mo2024} are trained using the proposed method, ensuring that every input image undergoes the proposed preprocessing step before entering the network. ConvNet is a conventional convolutional neural network architecture, without any specialized mechanisms for dealing with rotated inputs.

All experiments followed a consistent protocol: each model was trained and evaluated ten times using different random initializations. After training, we assess the accuracy of each configuration on the test set, where input images are systematically rotated from $0^\circ$ to $360^\circ$ in increments of $1^\circ$.

\begin{figure*}[t!]
    \centering
    % Primera fila: RotMNIST y CIFAR-10
    \begin{subfigure}[b]{0.49\textwidth}
        \centering
        \includegraphics[width=\textwidth]{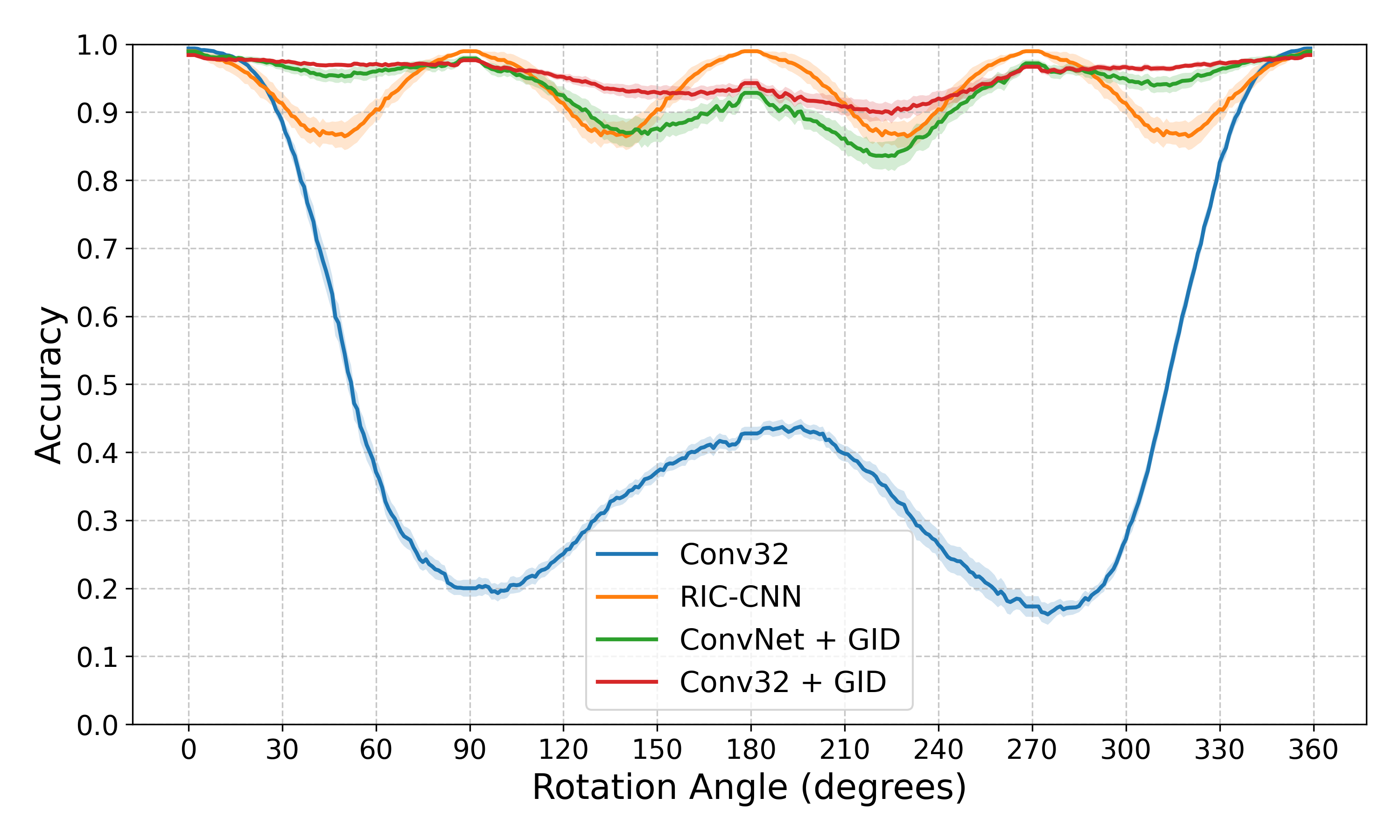}
        \caption{Rotated MNIST.}
        \label{fig:acc_360}
    \end{subfigure}
    \hfill
    \begin{subfigure}[b]{0.49\textwidth}
        \centering
        \includegraphics[width=\textwidth]{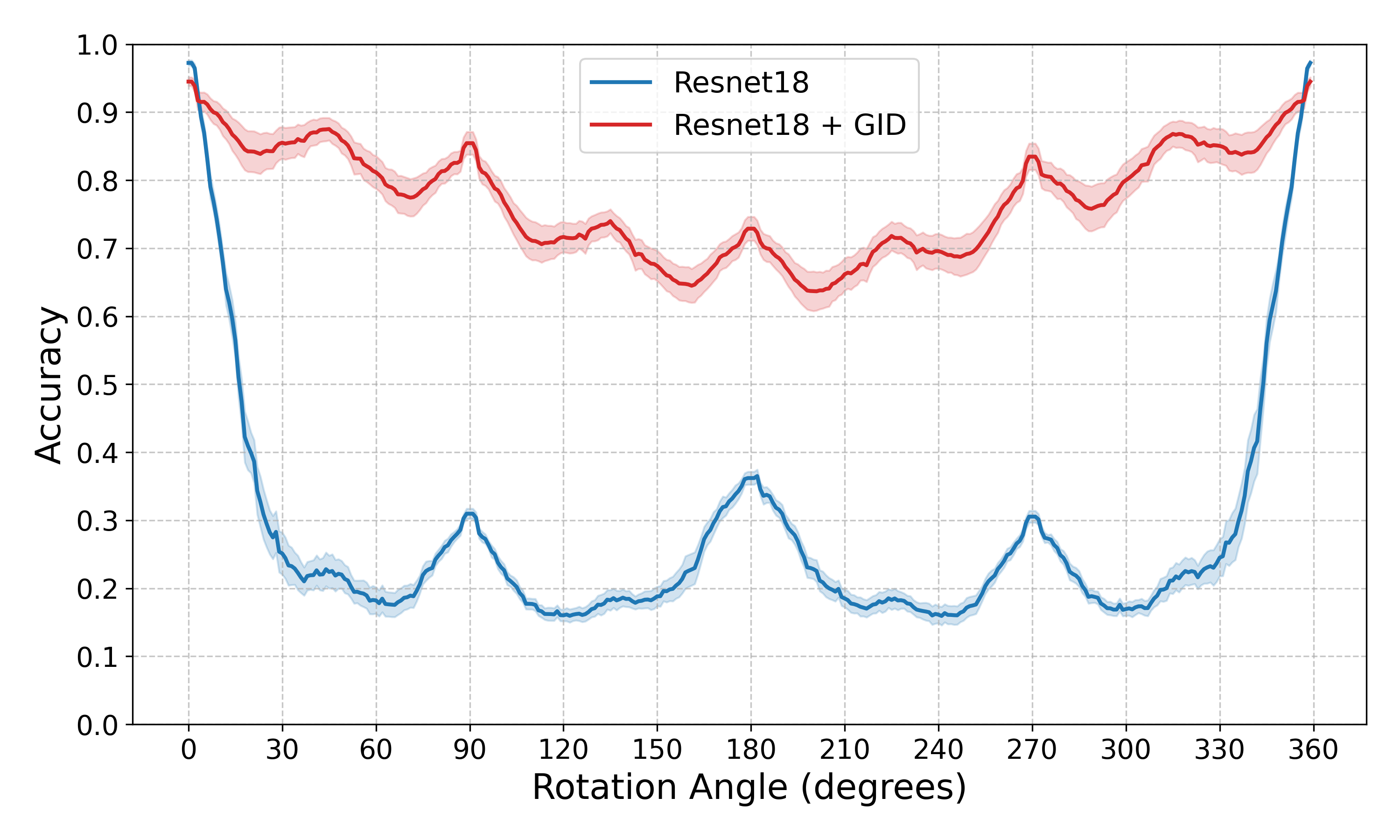}
        \caption{Rotated CIFAR-10.}
        \label{fig:cifar46}
    \end{subfigure}

    % Segunda fila: interpolación
    \begin{subfigure}[b]{0.49\textwidth}
        \centering
        \includegraphics[width=\textwidth]{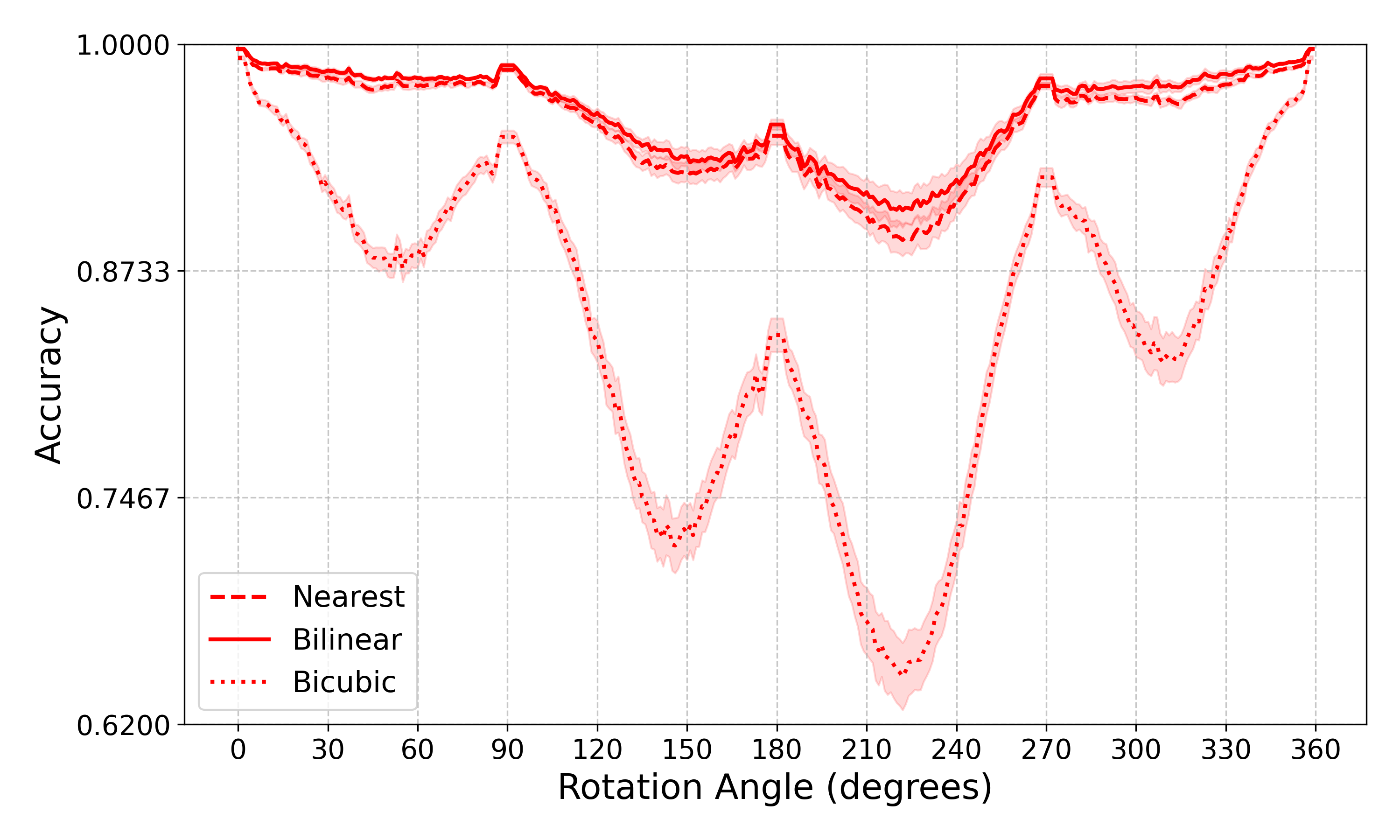}
        \caption{Interpolation sensitivity.}
        \label{fig:interpolation_comparison}
    \end{subfigure}

    \caption{Upper row: mean accuracy (with 10 repetitions) across rotation angles from $0^\circ$ to $359^\circ$ for different model configurations. (a) results on rotated MNIST; (b) results on rotated CIFAR-10. Lower row: (c) mean accuracy (with 10 repetitions) for GID+Conv32 using different interpolation methods, evaluated across all rotation angles in the rotated MNIST dataset. The shaded area represents one standard deviation.}
    \label{fig:rotation_curves_mnist_cifar}
\end{figure*}

To better understand the benefits of the proposed method, we compare the accuracy curves of these models against the baseline Conv32 model, which is trained without the proposed method, as well as against RIC-CNN~\cite{mo2024}, a rotation-invariant architecture specifically designed to improve performance on rotated data. The results of this comparison are illustrated in Figure~\ref{fig:acc_360}, where we observe that Conv32+GID achieves excellent performance, consistently maintaining high accuracy across the entire range of rotation angles.

To complement this analysis, we compare the classification accuracies of various models (Baseline CNN, ORN~\cite{orn2017}, RotEqNet~\cite{marcos2017}, G-CNN~\cite{cohen2016}, H-Net~\cite{worrall2017}, GA-CNN~\cite{hao2022}, B-CNN~\cite{delchevalerie2021}, E(2)-CNN~\cite{jenner2022}, DEF-CNN~\cite{dai2017}, RIC-CNN~\cite{mo2024}) on both the original MNIST test set and its rotated counterpart, RotMNIST, as reported in Table~\ref{tab:mnist_rotmnist_comparison}. This dataset is commonly used in previous works on rotation-invariant models, and using it allows for a direct comparison with existing approaches. In our case, RotMNIST is used exclusively for testing to assess the generalisation of the model to unseen orientations.

The results show that while most methods achieve similar performance on the original MNIST dataset (see corresponding column of Table~\ref{tab:mnist_rotmnist_comparison}), their accuracy differs significantly when evaluated on rotated images. In particular, we compare the test accuracy of ten trained instances of RIC-CNN and Conv32+GID. A t-test shows a statistically significant difference in favour of Conv32+GID, with a p-value of 0.0002 and a 99\% confidence interval for the mean accuracy difference. Both distributions pass the Shapiro–Wilk normality test at a 1\% significance level.

As shown in the rightmost column of Table~\ref{tab:mnist_rotmnist_comparison}, Conv32+GID achieves the highest accuracy on RotMNIST, confirming that the proposed method improves rotation robustness without architectural changes.

\begin{table*}[t!]
\caption{The classification accuracies from various methods on the original test set and rotated test sets of MNIST. Bold and underline stand for best and 2nd best results. Results for the comparison are obtained from \cite{mo2024}.}
\label{tab:mnist_rotmnist_comparison}
\begin{center}
\begin{tabular}{lccc}
\hline
Methods                         & Input size & MNIST & RotMNIST  \\ 
\hline
CNN                             & 32 x 32    & 99.55\%           & 45.42\%          \\
ORN \cite{orn2017}              & 32 x 32    & 99.42\%           & 80.01\%          \\
RotEqNet \cite{marcos2017}      & 28 x 28    & 99.26\%           & 73.20\%          \\
G-CNN \cite{cohen2016}          & 28 x 28    & 99.27\%           & 44.81\%          \\
H-Net \cite{worrall2017}        & 32 x 32    & 99.19\%           & 92.44\%          \\
GA-CNN \cite{hao2022}           & 28 x 28    & 95.67\%           & 93.29\%          \\
B-CNN \cite{delchevalerie2021}  & 28 x 28    & 97.40\%           & 88.29\%          \\
E(2)-CNN \cite{jenner2022}      & 29 x 29    & 98.14\%           & 94.37\%          \\ 
%\hline
DEF-CNN \cite{dai2017}          & 32 x 32    & \textbf{99.67\%}           & 46.97\%          \\
RIC-CNN \cite{mo2024}           & 32 x 32    & 99.02\%           & 95.52\%          \\
\hline
CONV32+GID                       & 32 x 32    & 98.58\%           & \textbf{96.32\%}          \\ 
\hline
\end{tabular}
\end{center}
\end{table*}

\subsection{Experiment 2: Sensibility to interpolation}
\label{subsec:experiment3}

Since the proposed method relies on rotating the input images to align them to a canonical orientation, interpolation becomes a critical factor in the overall pipeline. The way pixel values are computed during rotation can introduce artifacts or smooth variations that may influence the model’s performance, even when all other components remain unchanged. This experiment aims to assess how sensitive the method is to the choice of interpolation strategy in a controlled setting.

In this experiment, we evaluate the sensitivity of the proposed method to different interpolation strategies using the MNIST dataset. Specifically, we train GID+Conv32 using three interpolation methods: nearest-neighbour, bilinear, and bicubic. In each case, the interpolation is applied during the GID rotation step, while the rest of the network remains unchanged to ensure a fair comparison.

To ensure the reliability of the experiment, we repeat each configuration ten times with different random initializations. Once trained, we evaluate each model on the test set, where the input images are systematically rotated from $0^\circ$ to $360^\circ$ in increments of $1^\circ$, thereby covering the full range of orientations with uniform resolution.

We compare the accuracy curves obtained with each interpolation method across the full rotation range. The results of this comparison are shown in Figure~\ref{fig:interpolation_comparison}, which summarises the behaviour across angles. Among the three, bicubic interpolation yields the lowest accuracy, showing that the interpolation method used during alignment can influence the final performance, even when the architecture and training setup remain fixed.

\subsection{Experiment 3: Evaluation on CIFAR-10}
\label{subsec:experiment4}

In this experiment, we test the proposed method on natural images that include background information. We use the ResNet-18 architecture, specifically designed to handle the larger input size of the padded CIFAR-10 images. The model is trained using the same protocol described in Section~\ref{subsec:model_training}, ensuring that each input image passes through the proposed preprocessing step before entering the network.

To assess the robustness of the proposed method in this scenario, the model is trained and evaluated ten times using different random initializations. Each trained configuration is then tested on the CIFAR-10 test set, where the input images are systematically rotated from $0^\circ$ to $360^\circ$ in steps of $1^\circ$.

The impact of the preprocessing method is assessed by comparing the accuracy curve of ResNet-18 with GID against a baseline ResNet-18 model trained without any preprocessing. The results are presented in Figure~\ref{fig:cifar46}. We observe a clear improvement in accuracy across all rotation angles when the proposed method is applied. While the baseline model shows a significant drop in performance as the rotation angle increases, ResNet-18+GID maintains a considerably higher and more stable accuracy throughout the entire range.

\begin{figure*}[t!]
    \centering
    \begin{subfigure}[b]{0.47\textwidth}
        \centering
        \includegraphics[width=\textwidth]{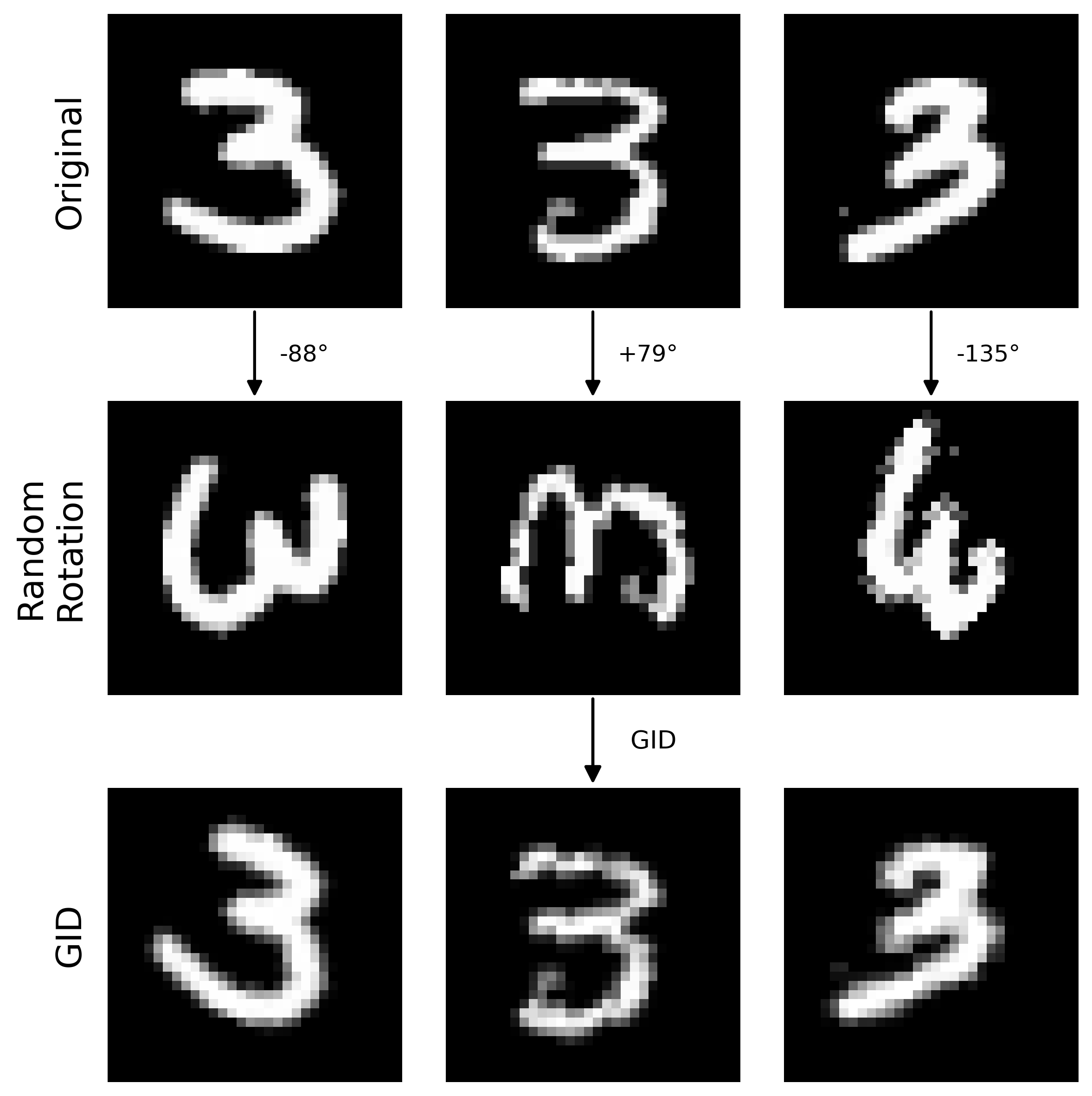}
        \caption{MNIST examples.}
        \label{fig:gid_vis_mnist}
    \end{subfigure}
    \hfill
    \begin{subfigure}[b]{0.43\textwidth}
        \centering
        \includegraphics[width=\textwidth]{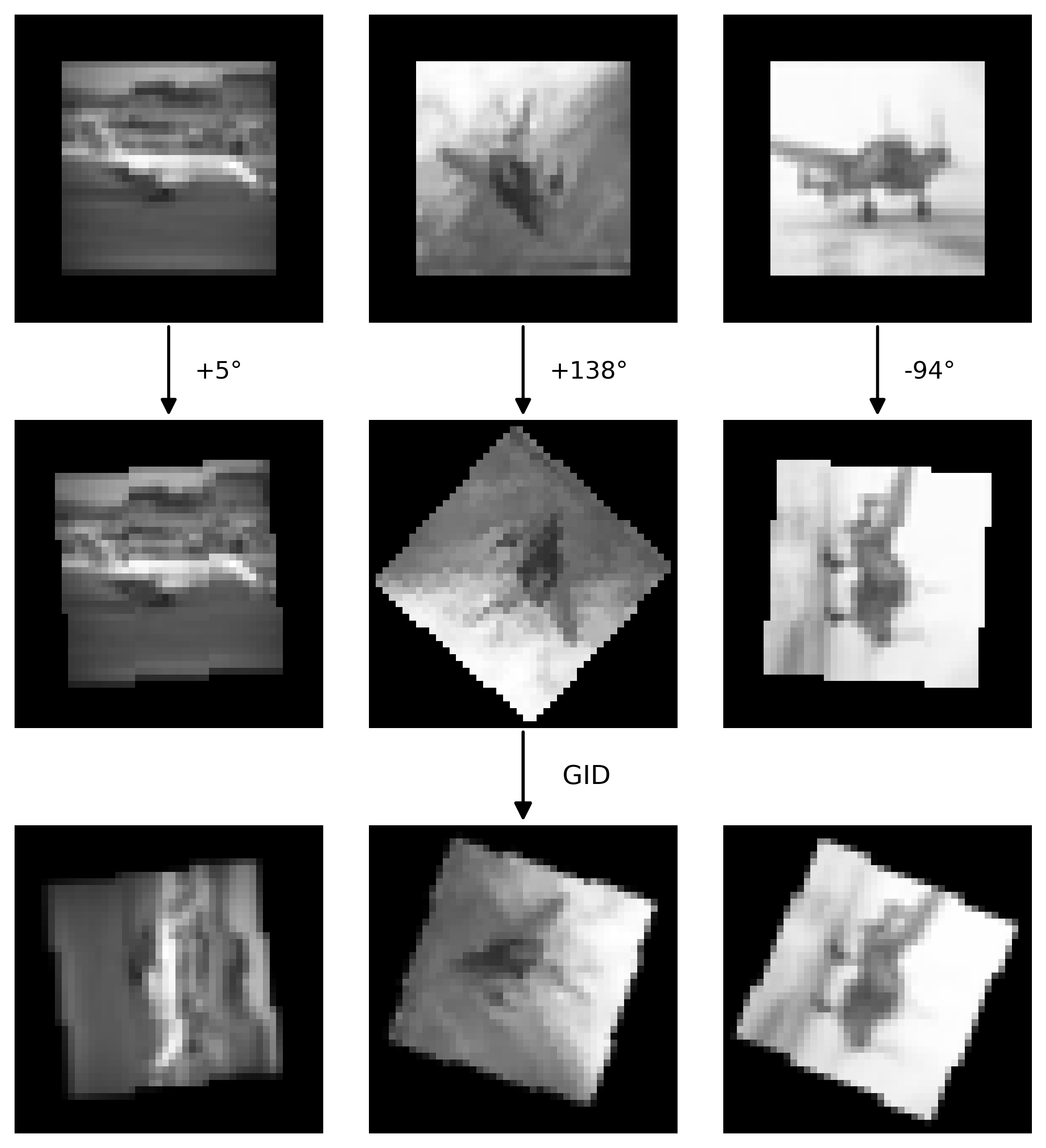}
        \caption{CIFAR-10 examples.}
        \label{fig:gid_vis_cifar}
    \end{subfigure}
    \caption{Effect of the GID preprocessing method on MNIST (a) and CIFAR-10 (b). In each case, the first row shows original images, the second row shows the same inputs after a random rotation, and the third row shows the result after applying GID. While GID successfully recovers the same orientation in MNIST, its performance is less reliable in CIFAR-10 due to background interference.}
    \label{fig:gid_visual_comparison}
\end{figure*}

This result confirms that the preprocessing step improves the model’s robustness to rotation, even when applied to more complex images with background, such as those in CIFAR-10. However, the gap with respect to the performance observed in MNIST is notable. One reason for this is that CIFAR-10 images include background information, which may interfere with the estimation of a consistent global orientation. For example, two images of the same object class might contain different backgrounds, leading the GID algorithm to compute different alignment angles despite belonging to the same category. This variability affects the consistency of the rotation alignment and limits the full potential of the method, as visually illustrated in Figure~\ref{fig:gid_visual_comparison}.

These findings reinforce the idea that the proposed preprocessing technique is best suited for cases where the object of interest is dominant or has been pre-segmented.

\section{\uppercase{Conclusions and future work}}
\label{sec:conclusions}

This paper introduced a preprocessing method that enhances rotation invariance in neural networks by estimating a global orientation for each image and rotating it to a canonical alignment. Unlike moment-based approaches that extract invariant descriptors, our method directly transforms the input while preserving spatial structure, making it compatible with standard convolutional networks and their usual processing pipelines.

In this work, we present a new preprocessing method that improves rotation robustness without requiring architectural modifications, demonstrating that such a simple step enhances model performance across rotated inputs. Experimental results on MNIST and RotMNIST show that our method improves classification accuracy over existing rotation-invariant architectures. The use of a black background in MNIST helps avoid interpolation artifacts during rotation, facilitating reliable alignment and preserving classification performance throughout the evaluation.

When applied to more complex datasets such as CIFAR-10, the method continues to provide improvements over baseline models. However, the presence of background elements introduces challenges, as these can affect the estimation of the global orientation during the alignment stage. This variability may lead to inconsistencies in the aligned outputs, especially when objects of the same class appear in different scenes or with varying context. The method proves most effective when the object of interest is centered and dominant in the image, which emphasizes the need for an additional preprocessing step to ensure that the object of interest is correctly segmented before applying the transformation.

% Future work will focus on addressing the method’s limitations for more diverse datasets. One important direction is the incorporation of a circular support for images, ensuring robustness to interpolation artifacts at the borders when applied to natural images. Another is the integration of object segmentation prior to alignment, improving orientation estimation in cluttered or unstructured scenes. Additionally, we aim to extend the evaluation to real-world scenarios where objects may not be centered, or where multiple objects appear in a single image, broadening the range of conditions under which the method is assessed.

% Future work will focus on addressing the method’s limitations for more diverse datasets. One important direction is the incorporation of a circular support for images, ensuring robustness to interpolation artifacts at the borders when applied to natural images. Another is the integration of object segmentation prior to alignment, improving orientation estimation in cluttered or unstructured scenes. Furthermore, we aim to evaluate the method's robustness under varying lighting conditions to address potential sensitivity to light source directions. Additionally, we aim to extend the evaluation to real-world scenarios where objects may not be centered, or where multiple objects appear in a single image, broadening the range of conditions under which the method is assessed.

Future work will address the method’s limitations across more diverse datasets. Key directions include incorporating circular support to ensure robustness against border interpolation artifacts in natural images and integrating object segmentation to improve orientation estimation in cluttered or unstructured scenes. Furthermore, we will evaluate the method's robustness under varying lighting conditions to address potential sensitivity to light source directions. Additionally, we will extend evaluations to real-world scenarios with uncentered or multiple objects, broadening the range of conditions under which the method is assessed.

\section*{\uppercase{Acknowledgements}}

C.V.A, M.M.G and E.S.S were funded by the Jaume I University through grant UJI-A2022-12. M.M.G and E.S.S were also funded by the Valencian Community Government through grant CIGE-2021-066.

\bibliographystyle{apalike}
{\small
\bibliography{example}}

\end{document}